\documentclass[10pt,twocolumn,letterpaper]{article}

\usepackage[pagenumbers]{cvpr} 

\usepackage{graphicx}
\usepackage{amsmath}
\usepackage{amssymb}
\usepackage{algorithm}
\usepackage{algpseudocode}
\usepackage{booktabs}
\usepackage{multirow}
\usepackage{graphics}
\usepackage{multicol}

\algdef{SE}[VARIABLES]{Variables}{EndVariables}
   {\algorithmicvariables}
   {\algorithmicend\ \algorithmicvariables}
\algnewcommand{\algorithmicvariables}{\textbf{hyperparameters}}

\usepackage[pagebackref,breaklinks,colorlinks]{hyperref}

\usepackage[capitalize]{cleveref}
\crefname{section}{Sec.}{Secs.}
\Crefname{section}{Section}{Sections}
\Crefname{table}{Table}{Tables}
\crefname{table}{Tab.}{Tabs.}

\def\cvprPaperID{40} 
\def\confName{CVPR}
\def\confYear{2023}

\begin{document}

\title{ColMix - A Simple Data Augmentation Framework to Improve Object Detector Performance and Robustness in Aerial Images}


\author{Cuong Ly\thanks{indicates equal contribution.}, Grayson Jorgenson\small{*}, \large{Dan Rosa de Jesus, Henry Kvinge, Adam Attarian, Yijing Watkins}\\
Pacific Northwest National Laboratory\\
902 Battelle Boulevard, Richland WA 99354\\
{\tt\small nhatcuong.ly@pnnl.gov}\\ 
{\tt\small firstname.lastname@pnnl.gov}
}
\maketitle

\begin{abstract}
In the last decade, Convolutional Neural Network (CNN) and transformer based object detectors have achieved high performance on a large variety of datasets. 
Though the majority of detection literature has developed this capability on datasets such as MS COCO, these detectors have still proven effective for remote sensing applications. Challenges in this particular domain, such as small numbers of annotated objects and low object density, hinder overall performance. In this work, we present a novel augmentation method, called {collage pasting}, for increasing the object density without a need for segmentation masks, thereby improving the detector performance. We demonstrate that collage pasting improves precision and recall beyond related methods, such as mosaic augmentation, and enables greater control of object density. However, we find that collage pasting is vulnerable to certain out-of-distribution shifts, such as image corruptions. To address this, we introduce two simple approaches for combining collage pasting with PixMix augmentation method, and refer to our combined techniques as {ColMix}. Through extensive experiments, we show that employing ColMix results in detectors with superior performance on aerial imagery datasets and robust to various corruptions.
\end{abstract}

\begin{figure*}[ht]
    \centering
    \includegraphics[scale=0.5]{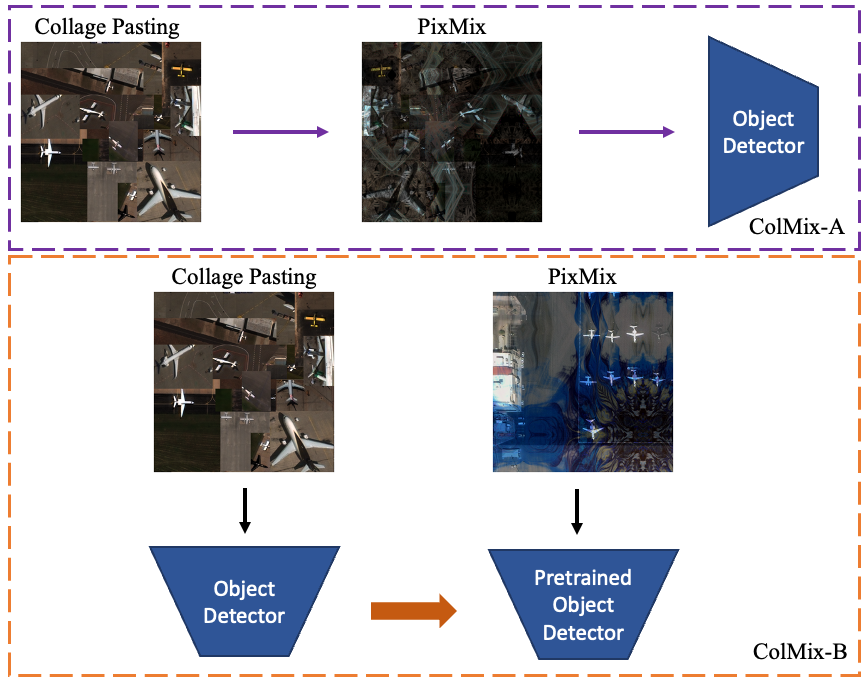}
    \caption{An overview of the proposed framework combining collage pasting and PixMix to improve model performance and robustness. In this study, we propose two simple approaches for combining collage pasting and PixMix. In the first approach, we combine them as a sequential processing pipeline, illustrated at the top of the figure. Each image represents the augmented image induced by the augmentation process specified at the top of that image. Meanwhile, we combine them in a hierarchical order in the second framework. Specifically, we pre-train an object detector using collage pasting and then fine-tune it further using PixMix. This process is shown at the bottom of the figure.}
    \label{fig:graphical_abstract}
\end{figure*}
\section{Introduction}\label{sec:intro}

Much of the existing work in object detection has been motivated by datasets such as MS COCO \cite{lin2014microsoft} and PASCAL VOC \cite{everingham2009pascal}. These datasets primarily consist of images that were taken by consumer cameras or cell phones. However, in domains such as aerial imagery, the data might be sourced from satellite image captures or cameras mounted on drones or aircraft, and this represents a significant distinction from datasets such as MS COCO.

In particular, in an aerial imagery dataset, there is often a greater prevalence of small objects, and the images tend to be higher resolution, thus, less densely populated. Further, the available annotation in an aerial imagery dataset is generally much smaller compared to natural image datasets, e.g., MS COCO, due to the prohibitive cost of data collection and annotation.

In this paper, we focus in part on the issue of object sparsity in aerial imagery datasets. For instance, in an aerial imagery such as the xView \cite{lam2018xview} dataset, on average less than $2\%$ of the image pixels are contained within target object bounding boxes. To what extent, if at all, do non-object background pixels contribute to improving model performance? Are many often essentially a waste of FLOPs? Is there an optimal object density for maximizing the performance from training?

A handful of data augmentation techniques have been developed to increase object density. Cutting and pasting objects from multiple training images into another with bounding boxes \cite{Hong2019} and segmentation masks \cite{Fang2019} \cite{Ghiasi2021}, or merging multiple images into one as in mosaic augmentation \cite{YOLOv3} are some of the current state-of-the-art methods in this area.

However, in aerial images, pixel-perfect segmentation masks can be an unaffordable luxury, and copying and pasting chips formed from just the bounding boxes of objects leads to image artifacts that can cause models to learn deleterious spurious correlations. And while simple and effective, mosaic augmentation provides only a limited amount of control over object density.

In this paper, we introduce a novel augmentation technique to address these issues, and refer to it as \emph{collage pasting}. This method enables fine-grained control of object density, avoids harmful image artifacts from pasting, and requires only bounding box annotations. The key idea is that instead of pasting tight bounding box chips of objects from other training images, we instead paste larger blocks, potentially containing multiple objects at a time and at random positions within the blocks. These copied regions are then randomly pasted within the target image to increase diversity. Mosaic augmentation can be seen as a special case of collage pasting, in which the blocks are tiled in an organized, non-overlapping manner for each training sample.\\
\indent Reducing the vulnerability of CNNs and transformer models to distribution shifts has been a major area of modern research. Image corruptions, in particular, are a distribution shift that has received significant attention: Hendrycks et al. \cite{corruptions}, and Michaelis et al. \cite{stylized} showed that CNNs perform poorly on test samples that contain blur, perceptible noise (Gaussian or shot noise), and natural distortion (e.g., snow) corruptions compared to unperturbed samples. Numerous studies \cite{corruption_aug,hendrycks2022pixmix,augmix,stylized,blur} have been designed to improve the model robustness on these common out-of-distribution shifts. Previous works such as \cite{corruption_aug} proposed including corruptions as a data augmentation process. Those frameworks have shown to be more resilient against the corruptions used as data augmentation during the training process and do not generalize well against unseen corruptions. For collage pasting, stitching larger blocks from multiple training images adds an unnatural transition between blocks. To a certain extent, using collage pasting during training can be seen as applying a type of corruption for data augmentation. Thus, using collage pasting can have a negative impact on the model's robustness against unseen corruptions. \\
\indent To improve the model robustness of collage pasting, we propose combining it with PixMix \cite{hendrycks2022pixmix} augmentation method, originally designed to improve model robustness against various corruptions for classification tasks.\\
\indent In particular, we introduce two simple yet effective techniques for combining PixMix with collage pasting to improve an object detection model's overall performance and robustness against corruptions. \Cref{fig:graphical_abstract} illustrates an overview of the proposed approaches. In the first approach, we combine collage pasting and PixMix in a sequential processing pipeline. Moreover, Reed et al. \cite{hierchial_pretraining} demonstrated that a model converges faster and provides more accurate overall performance when undergoing multiple pre-training stages, especially in a small dataset setting. Thus, exploiting that observation, we employ these two data augmentation methods in a hierarchical order in the second approach. Concretely, we pre-train an object detector with only collage pasting augmentation method and then finetune the object detector with only PixMix augmentation method. To the best of our knowledge, we are the first to combine data augmentations in such a framework for object detection in aerial imagery datasets. Through extensive experiments, we demonstrate that our proposed framework, named \textbf{ColMix}, consistently achieves the best performance on clean datasets while maintaining a competitive performance on corrupted datasets.

\section{Related Work}
\indent\textbf{Increasing object density via object cutting and pasting augmentation} has shown to be a simple and effective technique for improving the overall performance of an object detection model in aerial images. Hong et al. \cite{Hong2019} cut out objects from multiple training images based on their bounding boxes and directly pasted them onto another randomly selected training image regardless of background context. Using such an approach can easily cause overfitting due to the apparent artifact of discontinuity in semantic information around the objects, which an object detector might rely on to fit bounding boxes. Methods such as \cite{Fang2019} can optionally use context maps to avoid this problem, but doing so introduces higher computation costs. Meanwhile, Ghiasi et al. \cite{Ghiasi2021} used segmentation masks of objects to blend in the new background seamlessly. However, segmentation masks are not always available in many aerial imagery datasets. The mosaic augmentation technique, introduced with the YOLO family of detectors \cite{YOLOv3}, proposed combining multiple training samples by taking crops from each image and layering them side-by-side into a new training sample. The main limitation of the mosaic augmentation technique is the lack of fine-grained control over the resulting object density. In other words, the tiled chips can only be made so small and have a predictable structure to their arrangement that may lead to spurious correlations if made too small.

\textbf{Model robustness} ensures object detectors provide consistent performance regardless of the discrepancy between training and test distributions. For instance, object detectors should perform consistently under various weather conditions or input distortions. However, it is well-known that the performance of CNNs is susceptible to distribution shifts. To study model robustness, Hendrycks and Dietterich \cite{corruptions} evaluated CNN models on various datasets, each subjected to 15 corruptions at five levels of severity. A model is deemed robust when its performance does not deviate too much from the original clean dataset. In recent years, various studies \cite{corruption_aug,hendrycks2022pixmix,augmix,stylized,blur} have attempted to address model robustness. Studies such as \cite{corruption_aug} included a particular set of corruptions as a data augmentation method. They improved model robustness only under those corruptions seen during training and generalized poorly against unseen corruptions. Diverging away from using corruptions as data augmentation, Michaelis et al. \cite{stylized} proposed improving model robustness by training a model with the combination of the original dataset and its stylized version obtained from a neural style transfer model. However, the drawback of their work is the compromise in the performance of the original clean dataset. On the contrary, Hendrycks et al. \cite{hendrycks2022pixmix,augmix} designed data augmentation frameworks in separate studies to improve model robustness without compromising model performance on the original clean dataset. In the first study, Hendrycks et al. \cite{augmix} developed an augmentation method that mixes a combination of its augmented input, e.g., rotation, equalization, and shear, to generate a new sample. Meanwhile, they proposed to mix the inputs with structurally complex images from fractal image dataset \cite{fractal} and feature visualization of the initial layers \cite{feature_viz} in their second study \cite{hendrycks2022pixmix}.


\section{ColMix Augmentation}
\indent In this section, we detail our proposed framework. We first present collage pasting. Then, we present two approaches for incorporating PixMix with collage pasting to improve model robustness. 

\subsection{Collage Pasting Augmentation}

\begin{figure*}[ht]
    \centering
    \includegraphics[width=0.97\textwidth,height=0.65\textwidth]{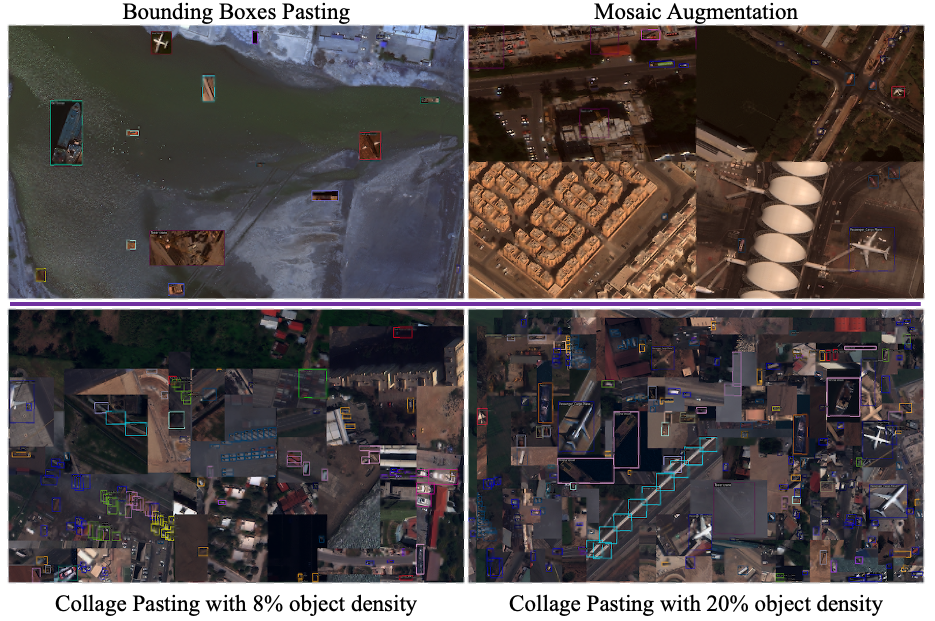}
    \caption{A comparison between bounding boxes pasting, mosaic augmentation, and collage pasting.}
    \label{fig:paste_comp}
\end{figure*}

Increasing object density by pasting multiple objects from different training samples has proven to be an effective way to improve an object detection model performance. However, some limitations still exist. For instance, pasting objects obtained directly from the bounding boxes creates a biased artifact that a model can use to learn to fit bounding boxes. On the other hand, mosaic augmentation avoids that pitfall, but it does not allow for fine-grained control over the object density. Hence, we introduce a new technique for pasting objects called collage pasting that facilitates fine-grained control over the object density while introducing fewer spurious artifacts. Concretely, we propose to paste larger blocks of multiple objects in each block from multiple training samples to create a new training sample. \Cref{fig:paste_comp} compares collage pasting and other object pasting methods. The figure shows that collage pasting can easily modulate object density compared to mosaic augmentation. In addition, pasting blocks containing more than the objects of interest alleviates pasting artifacts that a model can learn to depend on for fitting bounding boxes. 

This augmentation can be performed online during training, one sample at a time, or offline, generating an entire augmented dataset once before any training is done. Online augmentation increases the training time due to the extra cost of loading multiple training samples from which blocks are extracted. In this paper, we study the performance of the offline collage pasting method to reduce training times, i.e., generating a fixed number of unique epochs worth of augmented training data.

At a high level, the offline algorithm works by building upon either an existing training image or a blank image by breaking it down into a finite grid of ``corner positions'' and then iteratively pasting blocks copied from other samples at those positions. At each iteration, a single bounding box annotation is selected uniformly from the entire dataset and placed at a corner position that bounding boxes have not yet covered. The corresponding image is loaded into memory, and then the rectangular region of the selected bounding box is randomly expanded in all directions, subject to not overlapping existing bounding boxes in the target image beyond defined thresholds. After the expansion, any additional bounding boxes falling within the expanded region in the source image are included, and the new pixels are pasted into the target image. This process repeats until the desired object density has been reached. We provide precise details of the offline version in the pseudocode below, along with the following definition of variables used in the pseudocode:

\begin{itemize}
\setlength\itemsep{0em}
\item $TargetDensity$, objective object density,
\item $MinSize$, Minimum linear dimension of a pasted block, in pixels,
\item $MaxDilation$, Maximum expansion in any direction, in pixels,
\item $MaxExpansions$, Maximum number of expansion steps for a block,
\item $MinStep$, Minimum step size when expanding a block, in pixels,
\item $MaxStep$, Maximum step size when expanding a block, in pixels,
\item $OcclusionTol$, Maximum tolerated number of total pixels of overlap of a pasted block with existing bounding boxes,
\item $BBoxThreshold$, Used to remove bounding boxes from a paste source image with insufficient presence in the block, percent
\end{itemize}

\begin{algorithm*}
\begin{multicols}{2}
\begin{algorithmic}[1]
\State $\text{CornerPositions} \gets []$
\For{$0\leq i \leq \text{ImageWidth} // \text{MinSize}$}
\For{$0 \leq j \leq \text{ImageHeight} // \text{MinSize}$}
    \State $c\gets (i*\text{MinSize},j*\text{MinSize})$
    \If{$c$ is not contained in any bounding box}
        \State Append $c$ to CornerPositions
    \EndIf
\EndFor
\EndFor
\While{$\text{Sample object density} < \text{TargetDensity}$}
    \State Randomly select a bounding box annotation $b$ from \hspace*{1.2em} dataset,
    \State Load corresponding source image into memory,
    \State Rotate source image, random selection of \hspace*{1.2em} $\{0, 90, 180, 270\}$ degrees.
    \State Randomly shuffle CornerPositions
    \For{$c\in \text{CornerPositions}$}
        \State $\text{Margins} \gets [0,0,0,0]$
        \For{$0\leq i \leq \text{MaxExpansions}$}
            \State $\text{NewMargins}\gets\text{Margins}$
            \State $a \gets \text{Random}\{\text{MinStep},\ldots, \text{MaxStep})$
            \State Add $a$ to a random element of NewMargins, \hspace*{4.2em} replace by MaxDilation if the new sum is \hspace*{4.2em} larger.
            \State $B\gets$ the region of the target image of $b$ \hspace*{4.2em} expanded in each of the $4$ directions by \hspace*{4.2em} the elements of NewMargins, with top left \hspace*{4.2em} corner $x=c_0,y=c_1$.
            \If{$\text{\# of target image object pixels inside } B\leq\text{OcclusionTol}$}
                \If{$B$ is in image bounds}
                    \State $\text{Margins}\gets\text{NewMargins}$
                \EndIf
            \EndIf
        \EndFor
        \If{If at least one attempt was successful}
            \State Paste the pixels of the region from the \hspace*{4.2em} source image into $B$.
            \State Add all bounding boxes of the source image \hspace*{4.2em} at least BBoxThreshold\% contained in the \hspace*{4.2em} region to the target sample annotations.
            \State Exit loop
        \EndIf
    \If{no corner positions were successful}
        \State Paste just the bounding box at a random \hspace*{4.2em} corner position.
    \EndIf
    \EndFor
    \For{$c\in \text{CornerPositions}$}
        \If{$c$ is contained in a bounding box}
            \State Remove $c$ from CornerPositions
        \EndIf
    \EndFor
    \If{CornerPositions is Empty}
        \State Exit loop
    \EndIf
\EndWhile
\end{algorithmic}
\end{multicols}
\caption{Collage Pasting Pseudocode}
\label{alg:pseudo_code}
\end{algorithm*}

\subsection{Incorporating PixMix for Improving Model Robustness}
\indent Collage pasting is primarily designed to increase the object density in turn improving the detection performance of an object detector. However, training a model on collage pasting makes the model more susceptible to distribution shift. To a certain extent, collage pasting augmentation is a type of corruption. Using corruptions for data augmentation generalizes poorly to other types of corruptions, which has shown in \cite{corruption_aug}. Furthermore, machine safety is critical for successfully deploying a model in a real-world scenario. Therefore, the second objective of this work is to improve model robustness without compromising the performance of the original clean dataset.

We incorporate PixMix \cite{hendrycks2022pixmix} to provide model robustness in this work. PixMix is a data augmentation method in which inputs randomly blend with a subset of images chosen from fractals and feature visualizations. PixMix has consistently improved model performance across various machine learning safety measurements, such as corruptions, adversarial attacks, and anomaly detection while maintaining a solid performance on the original clean dataset for a classification task. In this study, we utilize PixMix as an additional augmentation method in two different ways. In the first approach, we include PixMix following collage pasting during training. Meanwhile, in the second approach, we utilize collage pasting and PixMix as a hierarchical pre-training process. Reed et al. \cite{hierchial_pretraining} showed that using a pre-trained model that has undergone multiple pre-training steps improves the overall performance. We use that observation to pre-train an object detector through collage pasting as an augmentation method. Then, we further finetune that object detector with PixMix. \\
\indent Moreover, the PixMix augmentation method used in this work is slightly different than previously proposed in \cite{hendrycks2022pixmix}. We use only images from the fractal image dataset for blending. In addition, we only blend the original input with images from the fractal image dataset and do not utilize the augmented version of the input in the blending process as described in \cite{hendrycks2022pixmix}. This customized PixMix was empirically determined. We refer to these two approaches as ColMix-A and ColMix-B, respectively, for brevity.


\section{Experiments \& Analyses}
\indent \textbf{Datasets} - We investigated the performance on two datasets: RarePlanes \cite{shermeyer2021rareplanes} and xView \cite{lam2018xview}. RarePlanes is a satellite imagery dataset containing real images captured with the WorldView-3 satellite and synthetic images. In this work, we only utilized real images portion of the dataset. It consists of more than 14000 annotated aircraft grouped into multiple categories. We leveraged three designated categories for the annotated aircraft: small, medium, and large civil aircraft, and used $512\times512$ resolution chips for the training and testing datasets, consisting of $5,567$ and $2,550$ samples, respectively. 

The xView \cite{lam2018xview} dataset contains hundreds of thousands of objects across 60 classes. However, for the experiments shown here, we use a smaller version of this dataset, which we call {xView-39}, that excludes classes such as the ``building'' and ``small car'' classes which dominate the others in terms of instance count \cite{lam2018xview}. We chose 39 classes, consisting of the vehicle, boat, aircraft, and construction equipment classes. Specifically, we formed xView-39 by excluding the following 21 classes from the original xView: ``Small car'', ``Flat car'', ``Tank car'', ``Straddle carrier'', ``Scraper/Tractor'', ``Ground grader'', ``Building'', ``Hut/Tent'', ``Shed'', ``Aircraft hanger'', ``Damaged building'', ``Facility'', ``Railway vehicle'', ``Helipad'', ``Pylon'', ``Shipping container'', ``Shipping container lot'', ``Storage tank'', ``Vehicle lot'', ``Construction site'', ``Tower structure''. In total, we had $4,002$ training samples and $894$ testing samples, each of $1333\times800$ resolution. 

\Cref{tab:collage_hyperparameters} provides the hyperparameters for generating the collage-augmented RarePlanes and xView-39 datasets used in our experiments.

\begin{table}[ht]
  \centering
  \resizebox{\columnwidth}{!}{
  \begin{tabular}{c|cc}
    \toprule
    Hyperparameter & RarePlanes & xView-39 \\
    \hline
    TargetDensity & $[0.05, 0.5]$  & $[0.01,0.3]$ \\
    MinSize & 25 & 25 \\
    MaxDilation & 512 & 1333\\
    MaxExpansions & 100 & 100 \\
    MinStep & 5 & 5 \\
    MaxStep & 30 & 30 \\
    OcclusionTol & 20 & 0\\
    BBoxThreshold & 50 & 50 \\
    \bottomrule
  \end{tabular}
  }
  \caption{Hyperparameter choices for generating the collage-augmented RarePlanes and xView-39 datasets for our experiments. The TargetDensity parameter was chosen uniformly from the specified intervals for each sample.}
  \label{tab:collage_hyperparameters}
\end{table}
\begin{table}[ht]
  \centering
  \resizebox{\columnwidth}{!}{ 
  \begin{tabular}{cc|cccc}
    \toprule
    \multirow{3}{*}{Model} & \multirow{3}{*}{Data Aug.} & \multicolumn{2}{c}{RarePlanes} & \multicolumn{2}{c}{xView-39} \\
    & & Clean & Corruptions & Clean & Corruptions\\
    & & mAP & mAPc & mAP & mAPc\\
    \hline
    \multirow{3}{*}{C-R50} & Baseline & 76.5 & 45.9 & 11.2 & 3.79 \\
    & ColMix-A & {79.5} & \textbf{56.3} & 13.9 & {5.63} \\
    & ColMix-B & \textbf{80.6} & 53.2 & \textbf{15.0} & \textbf{5.97}\\
    \hline
    \multirow{3}{*}{HTC-R50} & Baseline & 77.5 & 48.9 & 11.2 & 4.04 \\
    & ColMix-A & 80.3 & \textbf{57.6} & {14.5} & \textbf{5.93} \\
    & ColMix-B & \textbf{81.3} & 54.7 & \textbf{15.3} & {5.84}\\
    \hline
    \multirow{3}{*}{HTC-PVT} & Baseline & 79.8 & 56.7 & 14.9 & 6.13 \\
    & ColMix-A & {81.3} & 61.8 & \textbf{16.5} & \textbf{7.88} \\
    & ColMix-B & \textbf{82.6} & \textbf{64.0} & {16.0} & 7.18 \\
    \bottomrule
  \end{tabular}
  }
  \caption{The performance of various objection detection models with and without ColMix on the clean and corrupted RarePlanes and xView-39 datasets.}
  \label{tab:architecture_comp}
\end{table}
\begin{table*}[ht!]
  \centering
  \begin{tabular}{c|cccc}
    \toprule
    \multirow{3}{*}{Data Aug.} & \multicolumn{2}{c}{RarePlanes} & \multicolumn{2}{c}{xView-39} \\
    & Clean & Corruptions & Clean & Corruptions\\
    & mAP & mAPc & mAP & mAPc\\
    \hline
    Baseline & 76.5 & 45.9 & 11.2 & 3.79 \\
    \hline
    Mosaic Aug. \cite{YOLOv3} & 76.4 & 31.7 & 13.1 & 3.80 \\
    Copy\&Paste \cite{Ghiasi2021} & 75.9 & 40.8 & 10.8 & 3.30 \\
    \hline
    AugMix \cite{augmix} & 76.8 & 51.0 & 9.00 & 3.51 \\
    Stylized ($\alpha=1$) \cite{stylized} & 64.1 & 52.9 & 5.20 & 2.97 \\
    Org. + Stylized ($\alpha=1$) \cite{stylized} & 75.5 & \textbf{59.0} & 10.1 & 4.77 \\
    \hline
    ColMix-A & \underline{79.5} & \underline{56.3} & \underline{13.9} & \underline{5.63} \\
    ColMix-B & \textbf{80.6} & 53.2 & \textbf{15.0} & \textbf{5.97}\\
    \bottomrule
  \end{tabular}
  \caption{The performance of C-R50 when various data augmentation methods were applied during training process for RarePlanes and xView-39. The \textbf{bold} text indicates the best result, and the \underline{underline} text shows the second-best result.}
  \label{tab:comp}
\end{table*}

\indent \textbf{Corrupted Dataset} - To measure the model robustness, we evaluated the performance on the corrupted version of RarePlanes and xView-39. We perturbed test samples with 15 corruptions at five severity levels as described in \cite{corruptions}.

\indent \textbf{Metrics} - We employed mean average precision (mAP), which is the mean of the Average Precision metric computed over Intersection over Union (IoU) between 50\% and 95\% across all classes in a given dataset. For corrupted datasets, we measure the performance with mean Average Precision under all corruptions and severities:
\begin{equation}
    \text{mAPc} = \frac{1}{15}\sum_{c=1}^{15}\frac{1}{5}\sum_{s=1}^{5}\text{mAP}_{c,s}
\end{equation}

\indent \textbf{Implementation Details} - We implemented three object detectors with CNNs and transformers as backbone models. Specifically, we implemented Cascade-RCNN \cite{cascade_rcnn} with ResNet50 \cite{resnet} (C-R50) as backbone, and HTC \cite{htc} with ResNet50 (HTC-R50) and PVT \cite{pvt} (HTC-PVT) as backbone. We used the MS COCO pre-trained models and trained them with the SGD optimizer, a learning rate of 0.02, and a batch size of 16 for 12 epochs, except for the HTC-PVT model. The HTC-PVT model was pre-trained on ImageNet and trained with the Adam optimizer and a learning rate of 0.0001. We deployed the model on a single V100 GPU for training and evaluation. In the inference stage, no augmentation was applied to the test samples. 

\subsection{Effectiveness of ColMix Augmentation}
\indent We assessed the performance of the proposed framework on three object detectors to validate its generalization across different architectures, i.e., CNNs and transformers. \Cref{tab:architecture_comp} presents the performance of ColMix on clean and corrupted test sets. The results reported in these experiments are the mean of three separate runs. The table shows that ColMix effectively and significantly improves the model performance (at least +\textbf{1.5} in mAP) and robustness over the baseline across all settings. xView-39 is a much more challenging dataset due to its larger number of classes and variation in the object's size; thus, the overall result is much lower compared to RarePlanes dataset. 
Furthermore, ColMix substantially improves model robustness over the baseline, especially in the RarePlanes dataset.

\subsection{Comparisons with the State-of-the-Art}
\indent We compared our proposed ColMix with various data augmentation methods designed to improve object density or model robustness. For improving object density, we evaluated the performance of ColMix against mosaic augmentation (Mosaic Aug.) \cite{YOLOv3}, and Copy\&Paste \cite{Ghiasi2021}. For Mosaic Aug. \cite{YOLOv3}, we split an image into $i\times j$ tiles, where $i, j\in \{1,2,3,4\}$, and each tile from that image was randomly selected for stitching with other tiles from other samples to create a new augmented input.

On the other hand, we evaluated ColMix against image stylization \cite{stylized} and AugMix \cite{augmix} for model robustness comparison. The image stylization \cite{stylized} method uses a neural style transfer model to generate input samples with artistic style. We followed the same setting as described in \cite{stylized}. We reported the results when trained with only stylized images and combined the original and stylized images. The other method used in this comparison is AugMix \cite{augmix}. AugMix is similar to PixMix, with the main difference in that AugMix creates an augmented input sample by blending multiple augmented versions of itself, including altering annotation transformations (e.g., shear or rotation) instead of blending with images from fractals and feature visualizations. Furthermore, it utilizes Jensen-Shannon divergence (JSD) loss for regularization to prevent the model from memorizing the transformations inserted into the samples. For the experiments presented here, we incorporated AugMix without altering annotation transformations and JSD loss for a fair comparison since ColMix does not utilize either.

\Cref{tab:comp} presents the overall results of ColMix and the aforementioned methods on RarePlanes and xView-39. As seen from the table, ColMix achieves the best performance on the clean datasets and outperforms the others by a large margin. In addition, ColMix shows a competitive result on the corrupted datasets, and it does not perform as well as image stylization on the RarePlanes dataset. However, the trade-off for a much stronger performance of the image stylization on the corrupted dataset is the inferior performance on the clean dataset, which is undesirable.


\subsection{Ablation Study}\label{section:ab_study}

\begin{table}
  \centering
  \resizebox{\columnwidth}{!}{
  \begin{tabular}{c|cccc}
    \toprule
    \multirow{3}{*}{Data Aug.} & \multicolumn{2}{c}{RarePlanes} & \multicolumn{2}{c}{xView-39} \\
    & Clean & Corruptions & Clean & Corruptions\\
    & mAP & mAPc & mAP & mAPc\\
    \hline
    Baseline & 76.5 & 45.9 & 11.2 & 3.79\\
    \hline
    Collage Pasting & 79.3 & 40.3 & \underline{14.1} & 3.64\\
    PixMix & 77.1 & \textbf{57.2} & 10.8 & 4.67\\
    \hline
    ColMix-A & \underline{79.5} & \underline{56.3} & 13.9 & \underline{5.63} \\
    ColMix-B & \textbf{80.6} & 53.2 & \textbf{15.0} & \textbf{5.97}\\
    \bottomrule
  \end{tabular}
  }
  \caption{The performance comparison of collage pasting, PixMix, and ColMix on the RarePlanes and xView-39 datasets. The \textbf{bold} text indicates the best result, and the \underline{underline} text shows the second-best result.}
  \label{tab:ab_study}
\end{table}

\begin{figure*}[ht]
    \centering
    \includegraphics[width=0.95\textwidth,keepaspectratio]{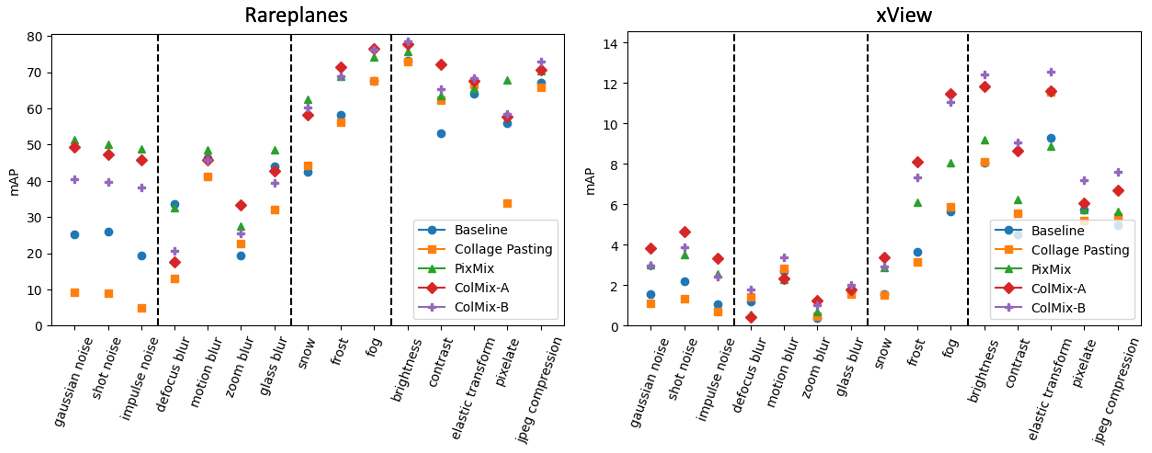}
    \caption{The mAP of C-R50 for each corruption when different augmentation method was used during training. Each point in the plot presents the mean mAP across five levels of severity.}
    \label{fig:corruptions_analysis}
\end{figure*}
\indent In this section, we investigated the contribution of each data augmentation method in ColMix to the overall performance. \Cref{tab:ab_study} presents results on RarePlanes and xView-39 datasets when only collage pasting, only PixMix, and ColMix are used as data augmentation. The table shows that collage pasting improves the overall performance on the clean datasets. However, its performance on the corrupted datasets deteriorates much worse than the baseline. Meanwhile, applying only PixMix allows for much higher performance on the corrupted datasets and a small fraction of improvement on the clean datasets over the baseline. Most importantly, by using ColMix, we achieve superior performance on the clean datasets while obtaining a robust performance. 

We then analyzed the effect of these augmentation methods on each corruption. \Cref{fig:corruptions_analysis} exhibits the mAP of C-R50 for each corruption used in this study when different augmentation methods were employed. Each point in the plot represents the mean mAP across five levels of severity. As seen in that figure, ColMix is resilient across all corruptions. Moreover, it performs as well as PixMix or even better in some cases. Meanwhile, the baseline model and collage pasting are extremely sensitive to most corruptions, especially noise corruptions, i.e., Gaussian noise, impulse noise, and shot noise.

\section{Conclusion}
\indent The challenges in detecting objects in aerial images are the small number of annotated objects and the sparsity of these objects in an image. In addition, it is well-known that CNNs are vulnerable to distribution shifts. Hence, we introduced ColMix, a new data augmentation framework, to improve model performance and robustness. ColMix provides a new technique for pasting objects without creating bounding box artifacts, allowing easy fine-grained control over the object density. Moreover, ColMix is more resilient to distribution shifts, e.g., corruptions, by incorporating PixMix in two simple yet effective approaches. This study shows that ColMix provides superior performance over other state-of-the-art augmentation methods across multiple datasets while obtaining competitive results against corruptions. Furthermore, ColMix can be easily extended to other tasks, e.g., instance segmentation, and other applications, such as detecting and counting cells in fluorescent microscopy images. 


{\small
\bibliographystyle{ieee_fullname}
\bibliography{eg}
}
\newpage

\end{document}